\documentclass{article}
\usepackage{lipsum}
\usepackage{graphicx} 
\usepackage{amsmath,amssymb, bm}
\usepackage[top=1in, bottom=1in, left=1in, right=1in]{geometry}
\usepackage{placeins}
\usepackage{xcolor}
\usepackage[utf8]{inputenc}
\usepackage{hyperref}
\hypersetup{
    colorlinks=true,
    linkcolor=blue,
    filecolor=blue,      
    urlcolor=blue,
    citecolor=blue
}
\usepackage{hhline}
\usepackage[round]{natbib}
\usepackage{booktabs}
\usepackage{colortbl}

\usepackage{algorithm}
\usepackage{algpseudocode}
\usepackage{listings}
\usepackage{xcolor}
\usepackage{amsmath}
\usepackage{multirow}
\usepackage{subcaption}

\definecolor{codegreen}{rgb}{0,0.6,0}
\definecolor{codegray}{rgb}{0.5,0.5,0.5}
\definecolor{codepurple}{rgb}{0.58,0,0.82}
\definecolor{backcolour}{rgb}{0.95,0.95,0.92}

\lstdefinestyle{mystyle}{
    backgroundcolor=\color{backcolour},   
    commentstyle=\color{codegreen},
    keywordstyle=\color{magenta},
    numberstyle=\tiny\color{white},
    stringstyle=\color{codepurple},
    basicstyle=\ttfamily\footnotesize,
    breakatwhitespace=false,         
    breaklines=true,                 
    captionpos=b,                    
    keepspaces=true,                 
    numbers=left,                    
    numbersep=5pt,                  
    showspaces=false,                
    showstringspaces=false,
    showtabs=false,                  
    tabsize=2
}

\definecolor{qqcustomcolor}{rgb}{0.3, 0.7, 0.6}

\definecolor{jzcustomcolor}{rgb}{0.36, 0.54, 0.66}  

\definecolor{stevencustomcolor}{rgb}{0.2 0.6 0.2}

\definecolor{sscustomcolor}{rgb}{0.7 0.4 0.3}

\title{Liger Kernel: Efficient Triton Kernels for LLM Training}
\author{Pin-Lun Hsu, Yun Dai, Vignesh Kothapalli, Qingquan Song, Shao Tang, \\ Siyu Zhu, Steven Shimizu, Shivam Sahni, Haowen Ning
 and Yanning Chen\\ \\
LinkedIn Inc}
\date{} 

\begin{document}

\maketitle

\begin{abstract}

Training Large Language Models (LLMs) efficiently at scale presents a formidable challenge, driven by their ever-increasing computational demands and the need for enhanced performance. In this work, we introduce \texttt{Liger-Kernel}, an open-sourced set of Triton kernels developed specifically for LLM training. With kernel optimization techniques like kernel operation fusing and input chunking, our kernels achieve on average 20\% increase in training throughput and a 60\% reduction in GPU memory for popular LLMs compared with HuggingFace implementations. In addition, \texttt{Liger-Kernel} is designed with modularity, accessibility and adaptability in mind, catering to casual and expert users. Comprehensive benchmarks and integration tests are built-in to ensure compatibility, performance, correctness and convergence across diverse computing environments and model architectures. 
The source code is available under a permissive license \href{https://github.com/linkedin/Liger-Kernel}{https://github.com/linkedin/Liger-Kernel}.
\end{abstract}

\section{Introduction}

Scaling Large Language Model (LLM)
training \citep{vaswani2017attention, wei2022emergent, brown2020language, team2023gemini, touvron2023llama, dubey2024llama} relies heavily on the stability of compute infrastructure and is susceptible to efficiency bottlenecks. 
Host/device memory management and latency-bandwidth trade-offs for tensor operations are central to the efficiency issues. However, beyond algorithmic scaling strategies, the true potential for optimization lies in fusing operations at the GPU kernel level, which minimizes memory copying and maximizes parallel efficiency. These last-mile kernel-level optimizations are crucial because any gains at this level are amplified by the inherent parallelism of GPUs, making them indispensable for improving overall training performance. Despite recent advancements in hardware and software usability for distributed training, optimizing the training process remains a highly complex and specialized task - which requiring
not only a deep understanding of both LLM algorithms and hardware architectures but also significant time and financial investments.

To address these challenges, we present \texttt{Liger-Kernel}, an open-source library of efficient Triton kernels \citep{tillet2019triton} for LLM training. \texttt{Liger-Kernel} enhances the efficiency and scalability of LLM training through a highly flexible and user-friendly interface. It streamlines complex tensor operations, minimizes computational overheads with kernel fusions~\citep{flashattention2022} and seamlessly integrates with diverse computing environments. Novice users can improve LLM training efficiency with a few lines of code, while advanced users can customize their model with modular components and adaptive layer configurations to suit their needs. \texttt{Liger-Kernel} requires minimal
dependencies, i.e., PyTorch~\citep{zhao2023pytorch}
 and Triton. \texttt{Liger-Kernel} supports multiple distributed frameworks such as PyTorch FSDP, DeepSpeed ZeRO~\citep{rasley2020deepspeed} and ZeRO++\citep{wang2023zero++,dai2024enhancing},  ensuring broad compatibility and performance optimization across various hardware platforms.

\section{Preliminaries}
\label{sec:preliminaries}
Eager mode execution in PyTorch \citep{paszke2019pytorch} provides a smooth development and debugging experience when authoring model code. However, step-by-step execution of PyTorch operations entails extra computational overheads, including function call stack, dispatching, and CUDA kernel launch latencies. In addition, materializing every intermediate activation for backward pass also introduces significant GPU memory usage. The majority of the efforts for addressing this issue have focused on model compilation and algorithmic operation fusion. Recently, more practitioners are implementing custom operation fusion in the Triton language \citep{tillet2019triton} to replace native PyTorch execution of model code.

\subsection{Model Compiler}

Model compilers transform high-level model descriptions (for example, \texttt{torch.nn.Module}) into optimized, low-level code that can be executed more efficiently, particularly on specialized hardware such as GPUs. Examples of such compilers include \texttt{torch.compile} \citep{pytorch2024}, TVM \citep{chen2018tvm}, XLA \citep{xla2020}, and nvFuser. \texttt{torch.compile} is the latest PyTorch-native model compilation feature introduced in PyTorch 2.0. Its frontend just-in-time (JIT) captures the computational graph and converts python-level operations into an intermediate representation (IR). Its backend performs low-level optimizations on the IR and translates into high-performance code in Triton  for GPUs and C++ with OpenMP for CPUs. Apache TVM provides a unified intermediate representation for various hardware platforms, aiming to bridge the gap between high-level deep learning frameworks and diverse deployment targets. XLA, developed by Google, is designed to optimize TensorFlow \citep{abadi2016tensorflow} and JAX \citep{frostig2018compiling} based training workflows. It performs operation fusion, layout optimization, and kernel generation tailored to the target hardware. nvFuser is a PyTorch-specific JIT compiler developed by NVIDIA. It is especially capable of generating optimized CUDA code tailored to the specific GPU, taking advantage of the GPU architecture's capabilities, such as memory hierarchy, parallelism, and instruction-level optimizations.

\subsection{An Algorithmic Perspective of Operation Fusion}
 The cornerstone of \texttt{Liger-Kernel}'s design is operation fusion. The main goal of the custom operation fusion is to mitigate the bottleneck arises between the high-bandwidth memory (HBM) and the shared memory (SRAM) for frequent memory copy. Each streaming multiprocessor (SM) needs fast access to data to execute multiple threads in parallel, but HBM, while large, is significantly slower than SRAM. This mismatch can lead to delays, where the processing cores sit idle, waiting for data to transfer from HBM to the faster, more limited SRAM. This becomes more severe in the context of deep learning models, especially those with large matrices (like in transformers) and numerous operations\footnote{ \cite{wen2022programming} provides more detailed strategies to alleviate this bottleneck and optimize GPU performance.}. Operation fusion combines several standalone GPU operations into a single one to avoid the per-op time and memory overhead in step-by-step execution mentioned at the beginning of Section \hyperref[sec:preliminaries]{2}. From an algorithmic perspective, operation fusion techniques like FlashAttention \citep{flashattention2022, flashattention22023} offer the advantage of optimizing specific computational patterns inherent to the algorithm itself, enabling more precise and tailored performance improvements compared to the broader, more generalized optimizations performed by model compilers. FlashAttention, for instance, optimizes the attention computation in transformer models by leveraging GPU memory hierarchies, reducing memory complexity from quadratic to linear. It splits the attention computation into smaller blocks that fit into the GPU on-chip SRAM, avoiding the need to materialize the full attention matrix and redundant memory accesses to the slower GPU high-bandwidth memory (HBM). FlashAttention-2 further improves this approach by reducing register spilling and enhancing parallelism across attention heads. These innovations collectively result in significant speedups and memory savings for attention computations, particularly for long sequence lengths.

\subsection{Custom Operation Fusion with Triton}
OpenAI's Triton is a programming language and compiler for high-performance GPU kernels with Python-like syntax (simpler than CUDA), making it easier to optimize deep learning operations without the complexity of low-level GPU programming. The JIT-compile nature of it also allows libraries and tools that use it to be more lightweight and portable. These features have increased the popularity of Triton for writing high-performance kernels for PyTorch on GPUs. \texttt{xFormers} \citep{xFormers2022} from Meta hosts interoperable and optimized Transformer building blocks implemented in Triton and CUDA and supports various attention mechanisms. he FlashAttention repository\footnote{\href{https://github.com/dao-ailab/flash-attention}{github.com/dao-ailab/flash-attention}}, in addition to hosting the CUDA implementation of FlashAttention algorithms, also includes other Transformer building block implementations (such as layer norm, a fused implementation of linear layer and squared ReLU activation etc) in Triton and \texttt{torch.script}. Unsloth\footnote{\href{https://github.com/unslothai/unsloth}{https://github.com/unslothai/unsloth}} from Unsloth AI re-implements popular LLMs \citep{touvron2023llama, mistral2023, phi2024} and LoRA \citep{lora2021} adapter layer in Triton to support efficient LLM fine-tuning and fast inference. Similar to the tiling design in FlashAttention, EfficientCrossEntropy\footnote{\href{https://github.com/mgmalek/efficient_cross_entropy}{https://github.com/mgmalek/efficient\_cross\_entropy}} fuses linear projection with CrossEntropy loss, and computes the loss in a block-wise manner to avoid materializing the entire logits tensor. \texttt{Liger-Kernel} draws inspiration and leverages code from some of the aforementioned projects as references. The details are presented in Section \hyperref[sec:kernels]{3.2}.

\section{Liger Kernel}

\subsection{API Design}

Ease of use is crucial for community adoption, and Liger kernels are designed to be accessible and straightforward. The guiding principle behind Liger's API design is to be the least disruptive to users' existing codebases while providing the flexibility needed for various levels of customization. Depending on the level of customization required, there are several ways to apply Liger kernels:

\begin{enumerate}
    \item \textbf{Using AutoLigerKernelForCausalLM:} The simplest way to leverage Liger kernels is through the \texttt{AutoLigerKernelForCausalLM} class. This approach requires no model-specific patching API imports. If the model type is supported, the modeling code will be automatically patched by Liger.

    \begin{lstlisting}[language=Python]
    from liger_kernel.transformers import AutoLigerKernelForCausalLM
    
    model = AutoLigerKernelForCausalLM.from_pretrained("path/to/some/model")
    \end{lstlisting}

    \item \textbf{Applying Model-Specific Patching APIs:} For fine-grained control over the model code, users can leverage \texttt{Liger-Kernel}'s model-specific patching APIs.  These APIs are versatile and can be used with various model architectures beyond causal language models, such as sequence classification.

    \begin{lstlisting}[language=Python]
    from liger_kernel.transformers import apply_liger_kernel_to_llama
    
    apply_liger_kernel_to_llama()
    model = AutoModelForSequenceClassification.from_pretrained("/path/to/some/model")
    \end{lstlisting}

    \item \textbf{Composing Custom Models:} Advanced users can leverage individual Liger kernels (as required) to create their own custom models. For instance, the torch-like code below illustrates the creation of a \texttt{LigerTransformer} module, which leverages \texttt{LigerLayerNorm} to implement the layer normalization functionality and \texttt{LigerCrossEntropyLoss} to create the loss function.

    \begin{lstlisting}[language=Python]
    import torch
    from liger_kernel.transformers import LigerLayerNorm, LigerCrossEntropyLoss

    class LigerTransformer(torch.nn.Module):
        def __init__(self, hidden_dim, *args, **kwargs):
            super().__init__()
            # create attn, mlp blocks or any custom operation
            ...
            # use Triton-optimized LigerLayerNorm
            self.layer_norm = LigerLayerNorm(hidden_dim)

        def forward(self, x):
            # forward pass of the model
            ...

    # use the Triton-optimized LigerCrossEntropyLoss
    loss_fn = LigerCrossEntropyLoss()
    \end{lstlisting}

\end{enumerate}

These flexible options ensure that Liger kernels can be easily integrated into various workflows, promoting efficient training and deployment of LLMs.

\subsection{Kernels}
\label{sec:kernels}

Throughout the discussion, vectors\footnote{Vectors are assumed to be column vectors unless otherwise specified.} and matrices are represented by bolded lowercase and uppercase letters, e.g., $\bm{x} \in \mathbb{R}^n$ and $\bm{\bm{W}}\in \mathbb{R}^{m \times n}$. The all-ones vector is denoted as $\bm{1}_n \in \mathbb{R}^n $. Functions are applied to the variable element-wise, i.e., $f(\bm{x})_i = f(x_i)$. We use $\odot$ to denote the element-wise product between tensors, and $^\top$ to denote the matrix transpose.  

In our kernel implementations, both input and output tensors are reshaped into two-dimensional matrices with the shape $(B \times T, H)$, where $B$ is the batch size, $T$ is the sequence length and $H$ is the hidden dimension. 

In each kernel, Triton parallelizes operations on each row of input\footnote{We compute the number of warps based on the block size, which is dependent upon the size of each row. We reuse the \texttt{calculate\_settings} function from \url{https://github.com/unslothai/unsloth/blob/main/unsloth/kernels/utils.py}.}.
Therefore, we focus on the mathematical operations given a row of input denoted as $\bm{x}$ and the corresponding output denoted as $\bm{y}$.  
In the backward pass, given a loss function $\mathcal{L}$, we use $\nabla_{\bm{y}}\mathcal{L}$ to denote the gradient back-propagated from $\mathcal{L}$ to $\bm{y}$.

\paragraph{RMSNorm.}

   We fuse the normalization and scaling steps of the RMSNorm computation into a single Triton kernel\footnote{The implementation is referenced the code from \url{https://github.com/unslothai/unsloth/blob/main/unsloth/kernels/rms_layernorm.py} and \url{https://triton-lang.org/main/getting-started/tutorials/05-layer-norm.html}.}. Specifically, given the input $\bm{x} \in \mathbb{R}^n$ and the learnable parameters $\bm{\gamma}\in \mathbb{R}^n$, the output $\bm{y} \in \mathbb{R}^n$ is defined as~\citep{zhang2019root}:
\begin{align}
\label{RMSNorm-forward}
    \bm{y} =  \hat{\bm{x}} \odot \bm{\gamma}, \hspace{20pt} \hat{\bm{x}} =  \frac{\bm{x}}{\textrm{RMS}(\bm{x})},
\end{align}
where $\hat{\bm{x}} \in \mathbb{R}^n$ is the normalized input, $\textrm{RMS}(\bm{x}) = \sqrt{\sum_i x_i^2/n + \epsilon}$ and $\epsilon$ is a small constant for numerical stability. In the backward pass, we have the gradient back-propagated to $\bm{x}$ and $\bm{\gamma}$ as
\begin{align} 
\begin{split}
    \nabla_{\bm{x}}\mathcal{L} &= \frac{1}{\textrm{RMS}(\bm{x})}\left(\nabla_{\bm{y}}\mathcal{L} \odot \bm{\gamma} - \underbrace{\left[\hat{\bm{x}}^\top(\nabla_{\bm{y}}\mathcal{L} \odot \bm{\gamma})/n \right]}_{\textrm{a numerical value}} \hat{\bm{x}}\right), \\
    \nabla_{\bm{\gamma}}\mathcal{L} &= \nabla_{\bm{y}}\mathcal{L} \odot \hat{\bm{x}}. \label{RMSNorm-backward}
\end{split}
\end{align}
Since the same ${\bm{\gamma}}$ is applied to all input vectors ${\bm{x}}$ in the same batch, the gradients need to be summed up.

\paragraph{LayerNorm.}
Similar to the RMSNorm, given the input $\bm{x} \in \mathbb{R}^n$, the learnable parameters $\bm{\gamma}\in \mathbb{R}^n$ and $\bm{\beta}\in \mathbb{R}^n$, the output $\bm{y} \in \mathbb{R}^n$ is defined as~\citep{ba2016layer}:
\begin{align}
\label{LayerNorm-forward}
    \bm{y} =  \Tilde{\bm{x}} \odot \bm{\gamma} + \bm{\beta}, \hspace{20pt} \Tilde{\bm{x}} =  \frac{\bm{x} - \Bar{\bm{x}}} {\textrm{RMS}(\bm{x} - \Bar{\bm{x}})},
\end{align}
where $\Tilde{\bm{x}} \in \mathbb{R}^n$ is the centered and normalized input, with $\Bar{\bm{x}} = \left(\sum_i x_i/n \right)\bm{1}_n$. In the backward pass, we have the gradient back-propagated to $\bm{x}$, $\bm{\gamma}$ and $\bm{\beta}$ as
\begin{align} 
\begin{split}
    \nabla_{\bm{x}}\mathcal{L} &= \frac{1}{\textrm{RMS}(\bm{x} - \Bar{\bm{x}})}\left(\nabla_{\bm{y}}\mathcal{L} \odot \bm{\gamma} - \underbrace{\left[\Tilde{\bm{x}}^\top(\nabla_{\bm{y}}\mathcal{L} \odot \bm{\gamma})/n \right]}_{\textrm{a numerical value}} \Tilde{\bm{x}} - \frac{1}{n} \left[(\nabla_{\bm{y}}\mathcal{L})^\top \bm{\gamma} \right] \bm{1} \right), \\
    \nabla_{\bm{\gamma}}\mathcal{L} &= \nabla_{\bm{y}}\mathcal{L} \odot \Tilde{\bm{x}} \\
    \nabla_{\bm{\beta}}\mathcal{L} &= \nabla_{\bm{y}}\mathcal{L}.   \label{LayerNorm-backward}
\end{split}
\end{align}
Since the same ${\bm{\gamma}}$ and ${\bm{\beta}}$ are applied to all input vectors ${\bm{x}}$ in a batch, the gradients need to be summed up\footnote{The efficient aggregation is non-trivial and three variants are benchmarked: plain aggregation in pytorch, two-stage aggregation from \url{https://github.com/Dao-AILab/flash-attention/blob/main/flash_attn/ops/triton/layer_norm.py} and atomic based aggregation in \url{https://triton-lang.org/main/getting-started/tutorials/05-layer-norm.html}. The latter two approaches perform much better than the vanilla aggregation and the second approach is currently adopted.}.

\paragraph{RoPE.} We fuse the query and key rotation embedding computation into a single kernel to reduce overheads. For each rotary position embedding computation, given the input $\bm{x} \in \mathbb{R}^d$, the token position $m$ and the rotation matrix $\bm{R}_{\Theta, m}^d \in \mathbb{R}^{d \times d}$, the output $\bm{y} \in \mathbb{R}^d$ is
\begin{align}
    \bm{y} =  \bm{R}_{\Theta, m}^d \bm{x}. \label{RoPE-forward}
\end{align}
Our implementation of RoPE assumes a rotation matrix in the form of HuggingFace model\footnote{\href{https://github.com/huggingface/transformers/blob/v4.44.2/src/transformers/models/llama/modeling\_llama.py\#L253}{https://github.com/huggingface/transformers/blob/v4.44.2/src/transformers/models/llama/modeling\_llama.py\#L253}} instead of the rotation matrix described in \citet{su2023enhanced}. Namely,
\begin{align*}
    \bm{R}_{\Theta, m}^d = 
    \begin{pmatrix}
\cos m \theta_1 & 0  & \dots & 0 & -\sin m \theta_1 & 0  & \dots & 0 \\
0 & \cos m \theta_2  & \dots & 0 & 0 & -\sin m \theta_2  & \dots & 0 \\
0 & 0  & \dots & 0 & 0 & 0  & \dots & 0 \\
\vdots & \vdots  & \ddots & \vdots & \vdots  & \vdots & \ddots & \vdots \\
0 & 0 & \dots & \cos m \theta_{d/2} & 0 & 0 & \dots  & -\sin m \theta_{d/2} \\
\sin m \theta_1 & 0  & \dots & 0 & \cos m \theta_1 &  0 & \dots & 0 \\
0 & \sin m \theta_2  & \dots & 0 & 0 & \cos m \theta_2  & \dots & 0 \\
0 & 0  & \dots & 0 & 0 & 0  & \dots & 0 \\
\vdots  & \vdots & \ddots & \vdots & \vdots  & \vdots & \ddots & \vdots \\
0 & 0 &  \dots & \sin m \theta_{d/2} & 0 & 0 &  \dots & \cos m \theta_{d/2}
\end{pmatrix}
\end{align*}
where the parameters $\Theta$ is model specific. 

In the backward pass, we have
\begin{align}
\label{RoPE-backward}
\nabla_{\bm{x}}\mathcal{L} = (\bm{R}_{\Theta, m}^d)^\top \nabla_{\bm{y}}\mathcal{L}.
\end{align}
In the implementation, due to the sparsity of $\bm{R}_{\Theta, m}^d$, we adopt the efficient computation in \cite{su2023enhanced}.

\paragraph{SwiGLU.} We fuse the element-wise operations in the SwiGLU computation into a single kernel. Given the input $\bm{x}\in \mathbb{R}^n$ and learnable parameters $\bm{\bm{W}}\in \mathbb{R}^{m \times n},\bm{V}\in \mathbb{R}^{m \times n},\bm{b}\in \mathbb{R}^m$ and $\bm{c}\in \mathbb{R}^m$, the output $\bm{y} \in \mathbb{R}^m$ is defined as~\citep{shazeer2020glu}:
\begin{align}
\begin{split}
    \bm{y} 
    &=\text{Swish}_{\beta=1}(\bm{W} \bm{x}+\bm{b})\odot(\bm{V} \bm{x}+\bm{c}) \\
    &=\text{SiLU}(\bm{W} \bm{x}+\bm{b})\odot(\bm{V} \bm{x}+\bm{c}),
\end{split}
\end{align}
where $\text{SiLU}(z) = z\sigma(z)$ and $\sigma(z) = (1+\textrm{exp}(-z))^{-1}$ is the sigmoid function. We only consider the $\beta=1$ case here where $\text{Swish}$ degenerates to \text{SiLU}, which aligns with the implementation of existing supported HuggingFace LLMs.  Denote the values $\bm{x_1} = \bm{W} \bm{x}+\bm{b} \in \mathbb{R}^m$ and $\bm{x_2} = \bm{V} \bm{x}+\bm{c} \in \mathbb{R}^m$, we implement the kernel to compute the forward pass as
\begin{align}
    \bm{y}(\bm{x_1}, \bm{x_2}) =\text{SiLU}(\bm{x_1})\odot \bm{x_2}. \label{SwiGLU-forward}
\end{align}
Recall $\nabla_{\bm{y}}\mathcal{L}$ as the gradient back-propagated from $\mathcal{L}$ to $\bm{y}$. In the backward pass, we have
\begin{align}
\begin{split}
    \nabla_{\bm{x_1}}\mathcal{L} &= \nabla_{\bm{y}}\mathcal{L} \odot \left[\sigma(\bm{x_1}) + \text{SiLU}(\bm{x_1})\odot(1 - \sigma(\bm{x_1}))\right]\odot \bm{x_2}, \\
    \nabla_{\bm{x_2}}\mathcal{L} &= \nabla_{\bm{y}}\mathcal{L} \odot \text{SiLU}(\bm{x_1}). \label{SwiGLU-backward}
\end{split}
\end{align}
\paragraph{GeGLU.} Similar to SwiGLU, we fuse the element-wise operations. Given the input $\bm{x}\in \mathbb{R}^n$ and learnable parameters $\bm{\bm{W}}\in \mathbb{R}^{m \times n},\bm{V}\in \mathbb{R}^{m \times n},\bm{b}\in \mathbb{R}^m$ and $\bm{c}\in \mathbb{R}^m$, the output $\bm{y} \in \mathbb{R}^m$ is defined as~\citep{shazeer2020glu}:
\begin{align}
    \bm{y} =\text{GELU}(\bm{W} x+\bm{b})\odot(\bm{V} x+\bm{c}),
\end{align}
where we use the \texttt{tanh} approximation of GELU \citep{hendrycks2016gaussian}. Formally,
\begin{align}
    \text{GELU}(z) \approx 0.5z\left(1+\tanh\left[\sqrt{2/\pi}\left(z+ 0.044715 z^3\right)\right]\right).
\end{align}
Similar to SwiGLU, denote the values $\bm{x_1} = \bm{W} \bm{x}+\bm{b} \in \mathbb{R}^m$ and $\bm{x_2} = \bm{V} \bm{x}+\bm{c} \in \mathbb{R}^m$. The forward pass can be computed as:
\begin{align}
    \bm{y}(\bm{x_1}, \bm{x_2}) =\text{GELU}(\bm{x_1})\odot \bm{x_2}. \label{GeGLU-forward}
\end{align}
In the backward pass, we have:
\begin{align}
\begin{split}
    \nabla_{\bm{x_1}}\mathcal{L} &= \nabla_{\bm{y}}\mathcal{L} \odot \nabla_{\bm{x_1}}\text{GELU}(\bm{x_1}) \odot \bm{x_2}, \\
    \nabla_{\bm{x_2}}\mathcal{L} &= \nabla_{\bm{y}}\mathcal{L} \odot \text{GELU}(\bm{x_1}), \label{GeGLU-backward}
\end{split}
\end{align}
where
\begin{align}
\begin{split}
\nabla_{\bm{x_1}}\text{GELU}(\bm{x_1}) \approx \, & 0.5 \odot \left(1+\tanh\left[\sqrt{2/\pi}\left(\bm{x_1}+ 0.044715 \bm{x_1}^3\right)\right]\right) \\
                               & + \sqrt{1/(2\pi)} \bm{x_1} \odot \left(1-\tanh^2\left[\sqrt{2/\pi}\left(\bm{x_1}+ 0.044715 \bm{x_1}^3\right)\right]\right)  \odot\left(1+0.134145\bm{x_1}^2\right).
\end{split}
\end{align}
\paragraph{CrossEntropy (CE).} We move the gradient computation to the forward function along with an inplace replacement of the logit tensor to avoid them being materialized simultaneously. We also adopt online softmax computation to compute the gradient on the fly. Given the input logits $\bm{x}\in \mathbb{R}^V$, where $V$ is the vocabulary size, and target one-hot encoded label $\bm{t}$, the output probabilities are given as:
\begin{align}
\bm{y} =\textrm{softmax}(\bm{x}), \label{CE-forward}
\end{align}
and the cross-entopy loss is defined as $\mathcal{L} = -\sum_i t_i \log (y_i)$. The gradient back-propagated to $\bm{x}$ is given by:
\begin{align}
\label{CE-backward}
\nabla_{\bm{x}}\mathcal{L} = \bm{y} - \bm{t}. 
\end{align}
Additionally, we also employ the safe $\log$ operation to avoid numerical instabilities.

\begin{figure}
    \centering
\includegraphics[width=0.9\linewidth]{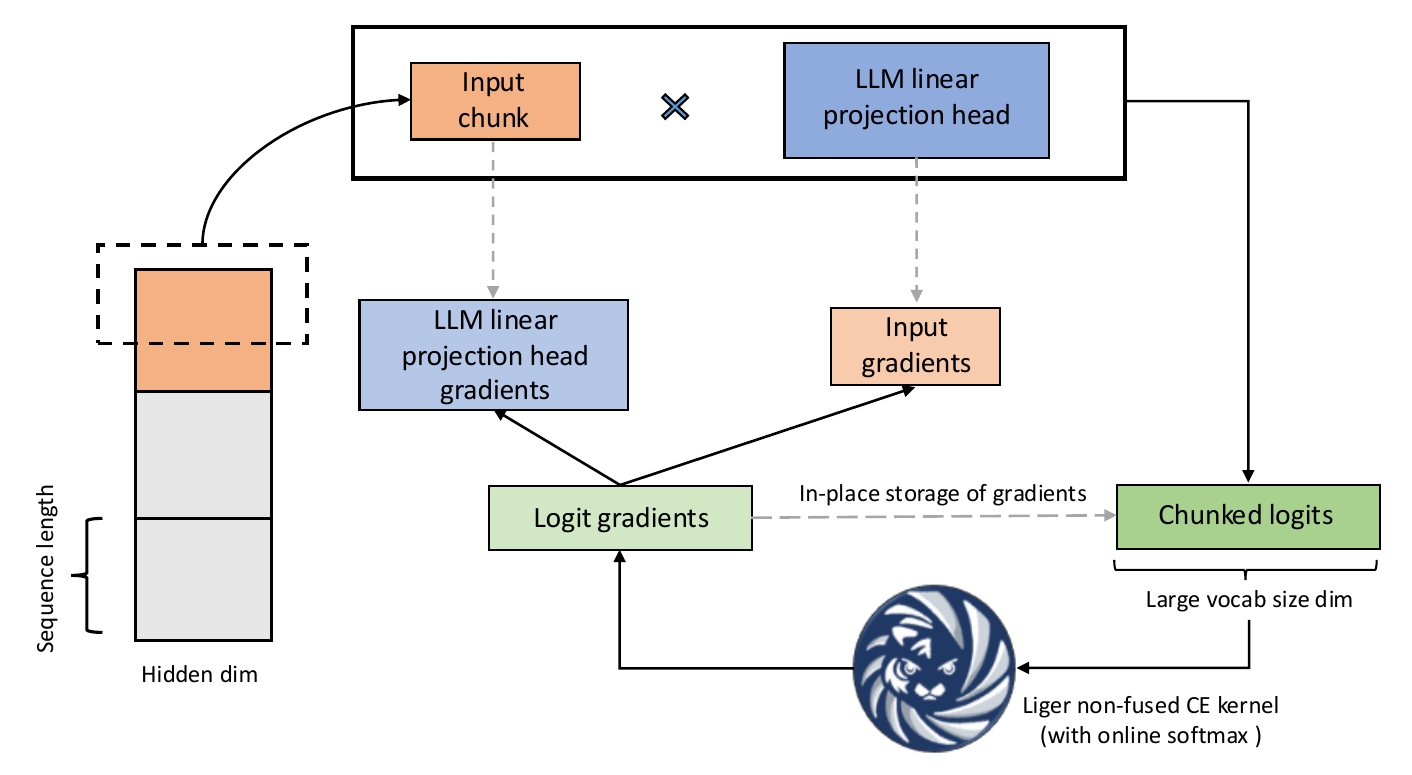}
    \caption{Fused Linear Cross Entropy.}
    \label{fig:flce}
\end{figure}

\paragraph{FusedLinearCrossEntropy (FLCE).} The rapid expansion of vocabulary in recent LLMs aims to enhance token granularity and achieve more compact prompt representations. However, this progress has revealed a significant challenge: the materialization of logit tensors during CE loss computation consumes excessive memory. This issue has become a major bottleneck in LLM training, limiting our ability to increase batch sizes and extend prompt contexts. Take the Gemma model as an example, single GPU training with a batch size of $8$ and sequence length of $4096$, the $256\textrm{k}$ vocabulary size will result in a $16.8$ GB logit tensor of precision bfloat16, causing a huge spike in the peak memory usage\footnote{The memory usually peaks at the end of each forward pass right before the release of the activations in the backward pass.}. 
Although the CE loss kernel considers an in-place replacement of gradient and logits, preventing the double materialization of two large tensors, single logit tensor size is still prohibitive in many cases which motivates us to explore the chunked logit and gradient computation to amortize the memory consumption\footnote{This is inspired from the GitHub discussions \url{https://github.com/pytorch/pytorch/issues/124480} and the solution from \url{https://github.com/mgmalek/efficient_cross_entropy}}. The main idea of FLCE is shown in Figure~\ref{fig:flce}. The 3D hidden states (shifted already to align with their next ground truth tokens) are flattened into a 2D matrix by collapsing the batch size and sequence length dimensions into a single dimension. The linear projection head is applied sequentially on the chunked hidden states. The generated output logits are passed to the non-fused Liger CE kernel to compute the partial loss and return the chunked logits gradient for deriving the chunked hidden states gradients and the accumulated projection head gradients.
\begin{align}
\begin{split}
&\bm{x}=\bm{W}^\top\bm{h}, \\
&\nabla_{\bm{h}}\mathcal{L} = \bm{W}\nabla_{\bm{x}}\mathcal{L}, \\
&\nabla_{\bm{W}}\mathcal{L} = \bm{h} (\nabla_{\bm{x}}\mathcal{L})^\top,
\label{FLCE-backward}
\end{split}
\end{align}

\noindent where $\bm{W}\in \mathbb{R}^{H \times V}$ denotes the linear projection head weight given vocabulary size $V$. $\bm{h} \in \mathbb{R}^{H}$ indicates a single row of the flattened hidden state matrix $\bm{H} \in \mathbb{R}^{BT \times H }$. A single row can be viewed as the special case with a chunk size equal to 1. $\bm{x}$ represents the logits projected from $\bm{h}$, for which, we have derived its gradient based on \eqref{CE-backward}. Since the same weight $\bm{W}$ is used for projecting all chunks, its final gradient needs to be summed up as $\nabla_{\bm{W}}\mathcal{L} = \sum_{\bm{h}} \bm{h} (\nabla_{\bm{x}}\mathcal{L})^\top$. Oftentimes, we can benefit from the compute-intensive behavior of the last layer projection, the overhead of block-wise matrix multiplications can be effectively compressed with delicate chunking on the tensor size to keep high GPU utilization with saturated operation time. In practice, we set the chunk size to be $ 2^{\lceil\log_2{\lceil\frac{BT}{\lceil V/H \rceil}\rceil}\rceil}$ with an intuition on picking the chunk size to be closer to the hidden dimension size to balance the trade-off between memory allocation and processing speed. 

\paragraph{Remark.} We additionally scale the gradients of the chunked inputs and the projection layer weights with the ratio of $\frac{\textrm{chunk size}}{B\times T}$. Formally, when a mean reduction is employed during the CrossEntropy loss calculation, the gradients are calculated for a particular input chunk and are not normalized over the entire input sequence. This additional scaling factor addresses such approximation issues.

\subsection{Testing Best Practices}

Testing is the cornerstone of our kernel development process. Exactness is non-negotiable, as even minor deviations can have far-reaching consequences. Through rigorous research and practical experience, we have distilled our approach into a set of best practices that ensure our kernels meet the highest standards of precision and reliability.

\subsubsection{Correctness}

Ensuring kernel precision is crucial, as any deviation from the original implementation could impact model convergence or cause critical errors. To achieve this, we prepare a pure PyTorch implementation (e.g., one provided by HuggingFace) for comparison and test the implementation with various input shapes and data types. We include regular shapes (e.g., powers of 2) and test irregular shapes to ensure proper handling of edge cases. We set appropriate absolute and relative tolerance levels: for fp32, use atol = $10^{-7}$ and rtol = $10^{-5}$; for bf16, use atol = $10^{-3}$ and rtol = $10^{-2}$ \footnote{Note that in practice, the tolerance may need further relaxation in some cases by one or two orders of magnitude, even for exact kernels. We use convergence tests to ensure exactness in cases where the tolerance for correctness needs to be loose.}.

Furthermore, large tensor dimensions can lead to inadvertent memory access issues. By default, the \texttt{program\_id} in the kernels are stored as \texttt{int32}. If \texttt{program\_id * Y\_stride > 2,147,483,647}, the value becomes negative, resulting in illegal memory access. Such overflows and incorrect memory addressing errors can be avoided by explicitly converting it to \texttt{int64} when dealing with large dimensions.

\subsubsection{Performance}

We ensure that the re-implementation of kernels in Triton is justified (compared to the baseline version) by testing across two key dimensions: speed and memory usage.

For input shapes in testing, we use actual dimensions/hyper-parameters from the training process, such as a batch size of $4$, a hidden dimension of $2048$, and a variable sequence length. This approach ensures that the test results reflect expected gains in production training across a family of models.

\subsubsection{Convergence Test}

In practical training settings, the contiguity, shape, and dtype of tensors might differ from the unit test conditions. To prove the validity of our computational gains, we mimic such real-world scenarios at a smaller scale and verify the exactness of logits, weights, and loss at the end of the training.

\subsubsection{Contiguity}

Since Triton operates directly on physical memory, non-contiguous tensors (where elements are not arranged sequentially) can lead to illegal memory access or incorrect outputs. For example, when deploying our RoPE kernel for production training, we observed significant loss divergence because the derivative from the \texttt{scaled\_dot\_product\_attention} function was not stored contiguously. To prevent such issues, it's best practice to ensure tensors are contiguous before passing them to the kernel.

\subsection{Integrations}
Liger has been successfully integrated with several popular training frameworks within the machine learning community, including Hugging Face transformers' \texttt{Trainer} class\footnote{\href{https://huggingface.co/docs/transformers/en/main_classes/trainer}{https://huggingface.co/docs/transformers/en/main\_classes/trainer}}, Hugging Face TRL's \texttt{SFTTrainer} class\footnote{\href{https://huggingface.co/docs/trl/main/en/sft_trainer}{https://huggingface.co/docs/trl/main/en/sft\_trainer}}, \texttt{Axolotl}\footnote{\href{https://axolotl-ai-cloud.github.io/axolotl/\#liger-kernel}{https://axolotl-ai-cloud.github.io/axolotl/\#liger-kernel}}, and \texttt{LLaMA-Factory}\footnote{\href{https://github.com/hiyouga/LLaMA-Factory}{https://github.com/hiyouga/LLaMA-Factory}}. These integrations demonstrate the flexibility and ease of use of the Liger API, enabling developers to leverage its optimization capabilities with minimal code changes. A simple flag is typically all that is needed to patch the model code with Liger kernels. For example:

\begin{lstlisting}[language=Python]
from trl import SFTConfig, SFTTrainer

trainer = SFTTrainer(
    "meta-llama/Meta-Llama-3-8B",
    train_dataset=dataset,
    # Setting `use_liger=True' will load the model using AutoLigerKernelForCausalLM
    args=SFTConfig(..., use_liger=True), 
)
trainer.train()
\end{lstlisting}

\section{Numerical Experiments}
This section presents the kernel level and end-end LLM training benchmarks using \texttt{Liger-Kernel} v0.2.1\footnote{\href{https://github.com/linkedin/Liger-Kernel/releases/tag/v0.2.1}{https://github.com/linkedin/Liger-Kernel/releases/tag/v0.2.1}}.

\subsection{Kernel Benchmark}

We benchmark the kernels individually across a variety of settings and illustrate the improvements in speed and memory consumption with Liger.

\paragraph{Setup.} All benchmarks are run on a single NVIDIA A100 GPU (80 GB). The CrossEntropy kernel is benchmarked on vocab sizes in the set $\{ 40960, 81920, 122880, 163840 \}$. The GeGLU and SwiGLU kernels are benchmarked on varying sequence lengths, whereas the RMSNorm, LayerNorm, and RoPE kernels are benchmarked on varying hidden dimensions. The sequence lengths and hidden dimension sizes are chosen from $\{ 4096, 8192, 12288, 16384 \}$. All benchmarks are repeated $10$ times to plot the median speed and memory along with $[0.2, 0.8]$ quantile values as the lower and upper bounds.

\begin{figure}
    \centering
    \begin{subfigure}{0.32\textwidth}
        \centering        \includegraphics[width=\textwidth]{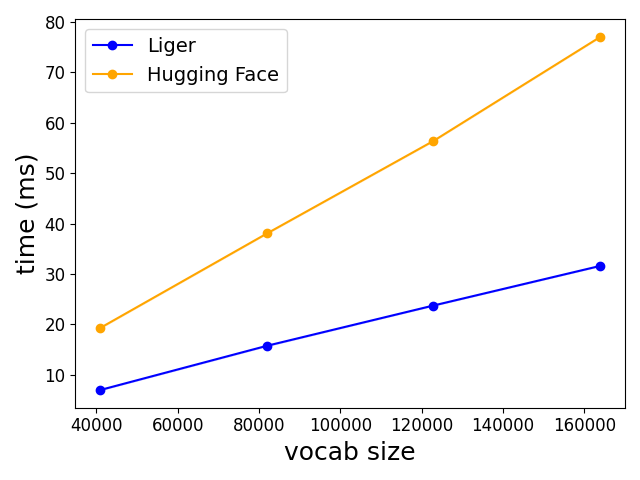}
        \caption{CrossEntropy}
        \label{fig:kernel_benchmarks_speed_ce}
    \end{subfigure}\hfill
    \begin{subfigure}{0.32\textwidth}
        \centering
        \includegraphics[width=\textwidth]{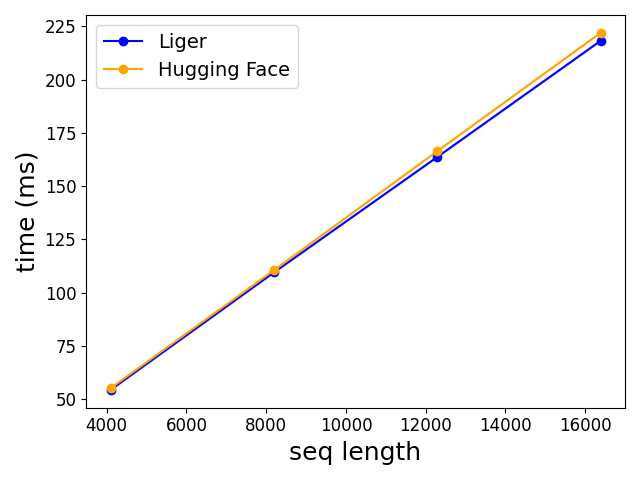}
         \caption{GeGLU}
         \label{fig:kernel_benchmarks_speed_geglu}
    \end{subfigure}
    \begin{subfigure}{0.32\textwidth}
        \centering
        \includegraphics[width=\textwidth]{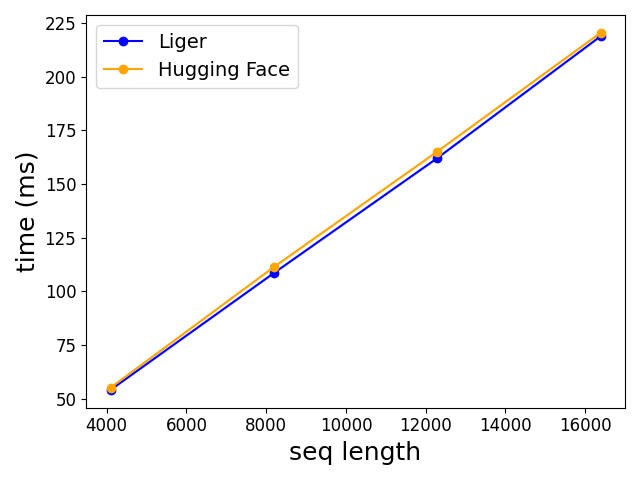}
         \caption{SwiGLU}
         \label{fig:kernel_benchmarks_speed_swiglu}
    \end{subfigure}
    \begin{subfigure}{0.32\textwidth}
        \centering
        \includegraphics[width=\textwidth]{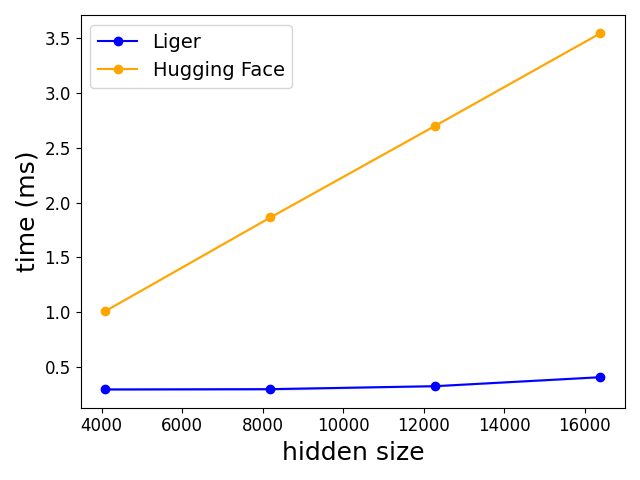}
         \caption{RMSNorm}
         \label{fig:kernel_benchmarks_speed_rmsnorm}
    \end{subfigure}
    \begin{subfigure}{0.32\textwidth}
        \centering
        \includegraphics[width=\textwidth]{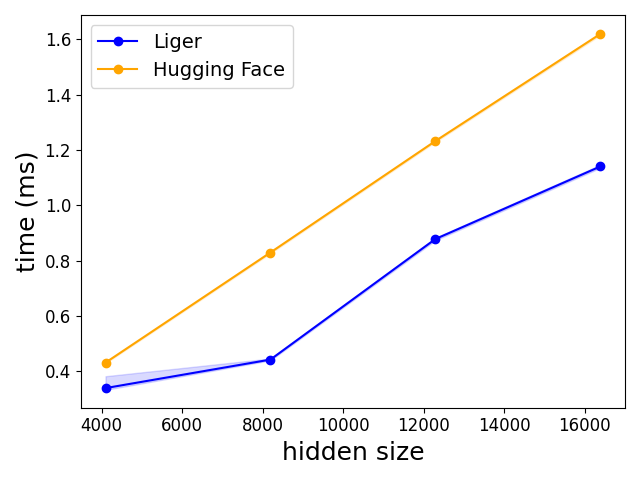}
         \caption{LayerNorm}
         \label{fig:kernel_benchmarks_speed_layernorm}
    \end{subfigure}
    \begin{subfigure}{0.32\textwidth}
        \centering
        \includegraphics[width=\textwidth]{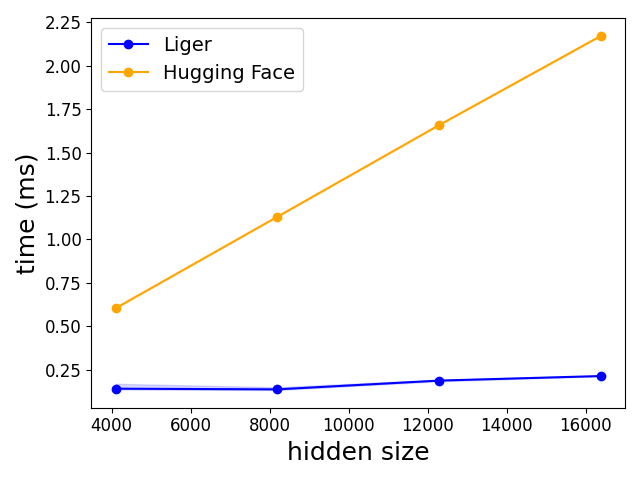}
         \caption{RoPE}
         \label{fig:kernel_benchmarks_speed_rope}
    \end{subfigure}
    \caption{Kernel execution speed benchmarks.}
    \label{fig:kernel_benchmarks_speed}
\end{figure}

\begin{figure}
    \centering
    \begin{subfigure}{0.32\textwidth}
        \centering
        \includegraphics[width=\textwidth]{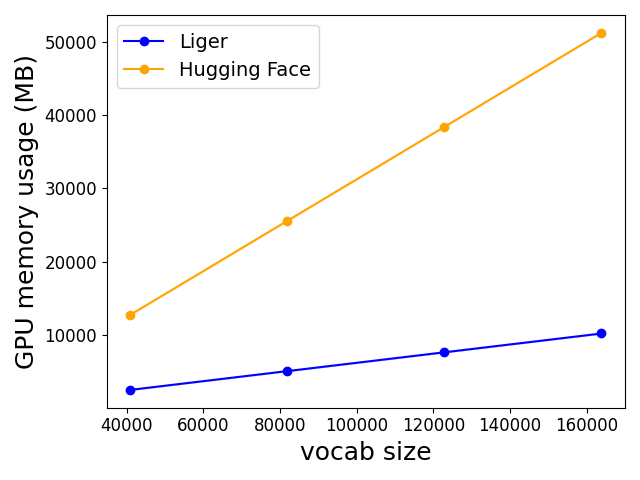}
        \caption{CrossEntropy}
        \label{fig:kernel_benchmarks_mem_ce}
    \end{subfigure}\hfill
    \begin{subfigure}{0.32\textwidth}
        \centering
        \includegraphics[width=\textwidth]{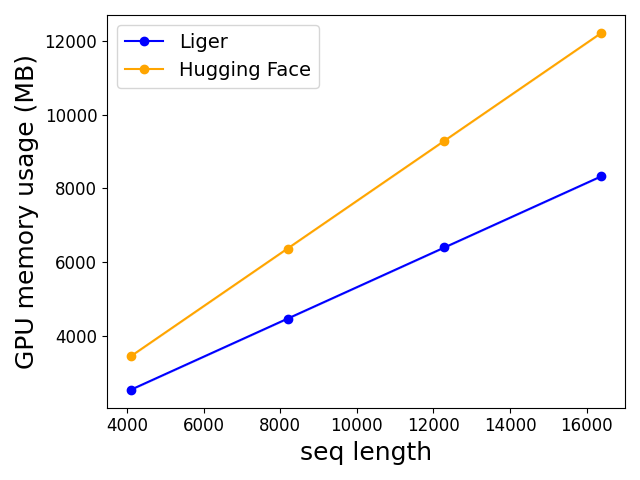}
         \caption{GeGLU}
         \label{fig:kernel_benchmarks_mem_geglu}
    \end{subfigure}
    \begin{subfigure}{0.32\textwidth}
        \centering
        \includegraphics[width=\textwidth]{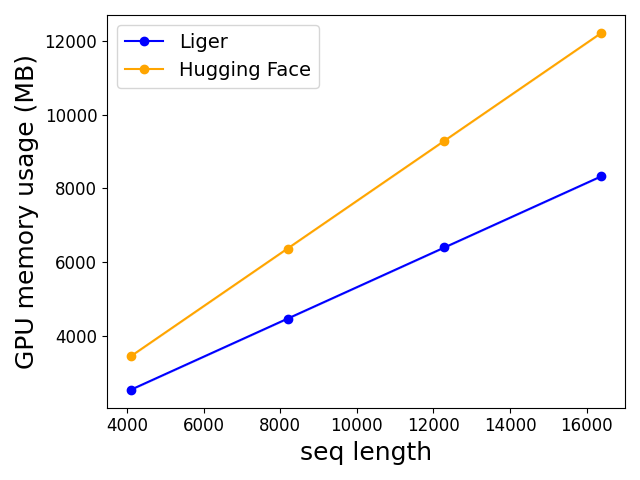}
         \caption{SwiGLU}
         \label{fig:kernel_benchmarks_mem_swiglu}
    \end{subfigure}
    \begin{subfigure}{0.32\textwidth}
        \centering
        \includegraphics[width=\textwidth]{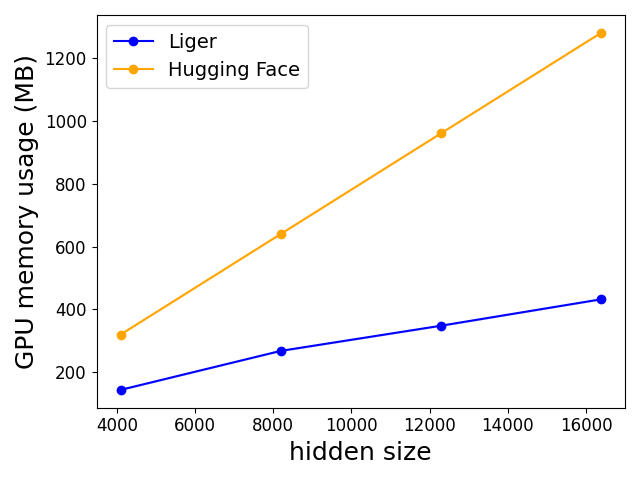}
         \caption{RMSNorm}
         \label{fig:kernel_benchmarks_mem_rmsnorm}
    \end{subfigure}
    \begin{subfigure}{0.32\textwidth}
        \centering
        \includegraphics[width=\textwidth]{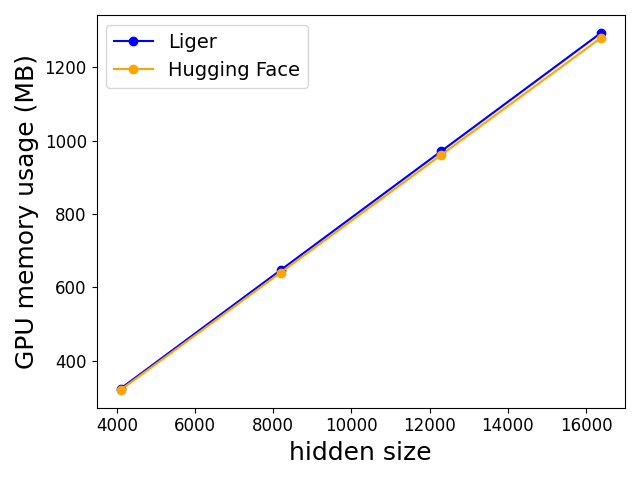}
         \caption{LayerNorm}
         \label{fig:kernel_benchmarks_mem_layernorm}
    \end{subfigure}
    \begin{subfigure}{0.32\textwidth}
        \centering
        \includegraphics[width=\textwidth]{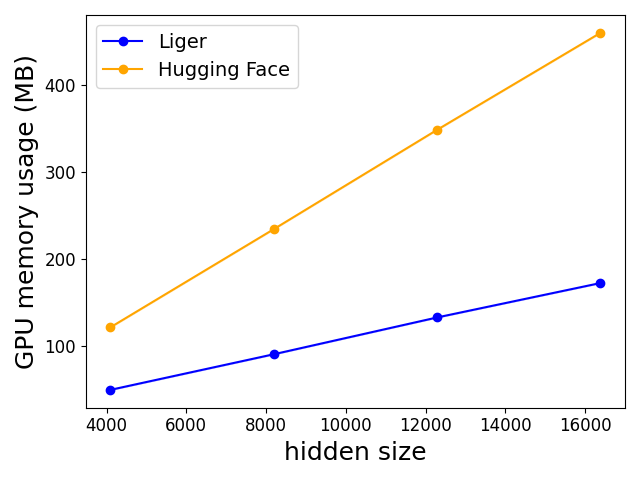}
         \caption{RoPE}
         \label{fig:kernel_benchmarks_mem_rope}
    \end{subfigure}
    \caption{Kernel peak allocated memory benchmarks.}
    \label{fig:kernel_benchmarks_mem}
\end{figure}

\paragraph{Results.} The kernel speed and memory benchmarks are illustrated in Figure \ref{fig:kernel_benchmarks_speed}, \ref{fig:kernel_benchmarks_mem} respectively. Observe that all the Liger-kernel implementations either execute faster, consume less memory or provide both of these benefits when compared to the baseline implementations. In the case of the CrossEntropy kernel, the online softmax computation along with in-place replacement of the kernel inputs with their gradients leads to approximately $ 3\times$ faster execution (Figure \ref{fig:kernel_benchmarks_speed_ce}) and consumes approximately $ 5\times$ less memory (Figure \ref{fig:kernel_benchmarks_mem_ce}) for a vocab size of $163840$. For GeGLU and SwiGLU, we maintain parity with the baseline in terms of speed (Figure \ref{fig:kernel_benchmarks_speed_geglu}, \ref{fig:kernel_benchmarks_speed_swiglu}) and reduce the peak memory consumption by roughly $ 1.6\times$ (when sequence length is $16384$) by recomputing the SiLU$(\cdot)$ and GELU$(\cdot)$ outputs during the backward pass (Figure \ref{fig:kernel_benchmarks_mem_geglu}, \ref{fig:kernel_benchmarks_mem_swiglu}). 

The RMSNorm implementation fuses the normalization and scaling operations into a single triton kernel and caches the root mean square values for usage in the backward pass. This avoids repetitive data transfers and floating point operations with minimal memory overheads. Figure \ref{fig:kernel_benchmarks_speed_rmsnorm}, \ref{fig:kernel_benchmarks_mem_rmsnorm} illustrates approximately $ 7\times$ reduction in execution time and roughly $ 3\times$ reduction in peak memory consumption for a hidden dimension of $16384$ respectively. A similar caching approach for the inverse root mean square is employed for LayerNorm kernel which results in approximately $ 30\%$ reduction in execution time (Figure \ref{fig:kernel_benchmarks_speed_layernorm}) with minimal memory overheads (Figure  \ref{fig:kernel_benchmarks_mem_layernorm}). Finally, for the RoPE kernel, we employ a flattened 1D tensor to represent the rotation matrix and leverage the repeated blocks in $\bm{R}_{\Theta, m}^d$ to significantly reduce the growth in latency with an increase in hidden dimension size. In particular, we achieve approximately $ 8\times$ speedup with approximately $ 3\times$ lower memory consumption for a hidden size of $16384$.

\subsection{Usecase Benchmark}
\paragraph{Setup.}  For the end-end training experiments, we employ $4$ NVIDIA A100 GPUs ($80$ GB each) to fine-tune the LLMs (LLaMA 3-8B, Qwen2, Gemma, Mistral, and Phi3) on the Alpaca dataset. We vary the batch size, set the precision to bfloat16, and use the AdamW optimizer with a cosine learning rate scheduler. The sequence length for training is set to $512$ tokens. The throughput and GPU memory usage metrics are collected after $20$ training steps with the standard error measured from $5$ repetitive runs. The benchmark script can be found in our GitHub repository\footnote{\href{https://github.com/linkedin/Liger-Kernel/tree/main/examples/huggingface}{https://github.com/linkedin/Liger-Kernel/tree/main/examples/huggingface}}. 


\paragraph{Performance Comparison.} 

\begin{figure}[H]
    \centering
    \begin{minipage}{0.45\textwidth}
        \centering
        \includegraphics[width=\textwidth]{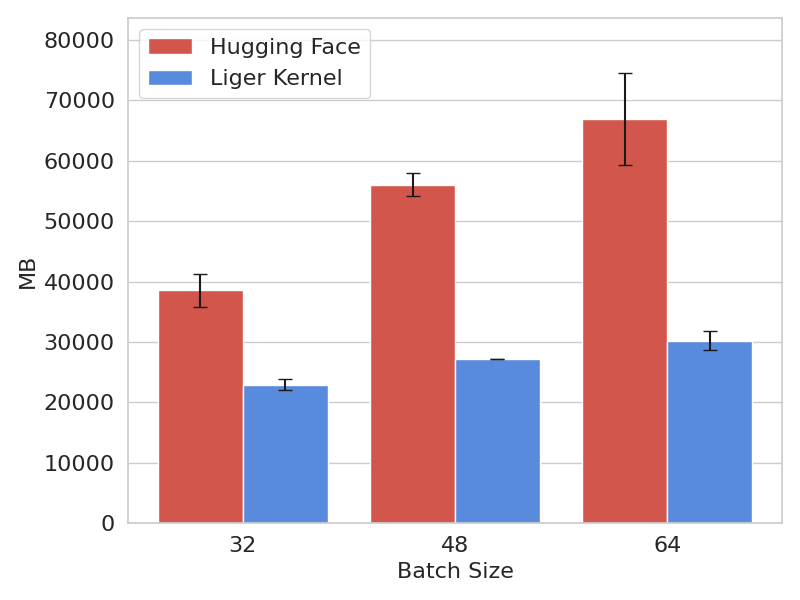}
        \label{fig:memory_llama3_8b}
    \end{minipage}\hfill
    \begin{minipage}{0.45\textwidth}
        \centering
        \includegraphics[width=\textwidth]{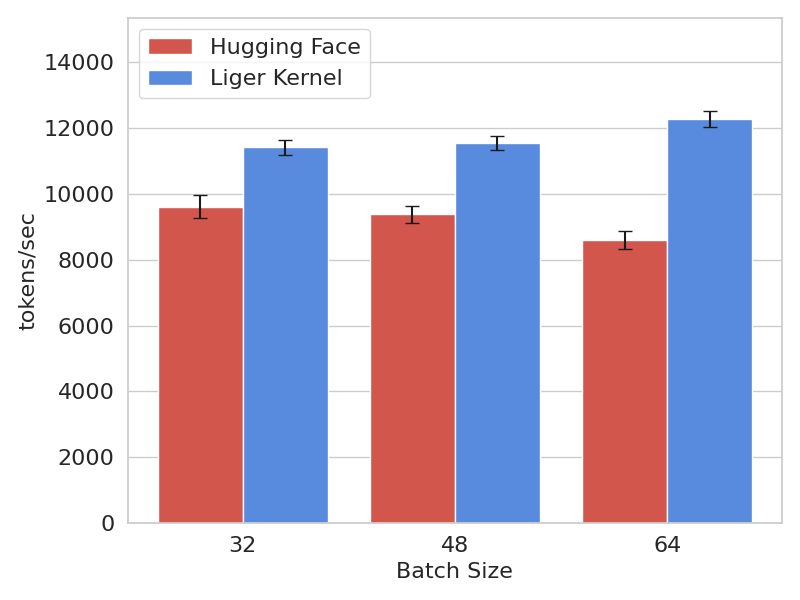}
        \label{fig:throughput_llama3_8b}
    \end{minipage}
    \caption{Comparison of peak allocated memory and throughput for LLaMA 3-8B.}
    \label{fig:combined_llama3_8b}
\end{figure}

\begin{figure}[H]
    \centering
    \begin{minipage}{0.45\textwidth}
        \centering
        \includegraphics[width=\textwidth]{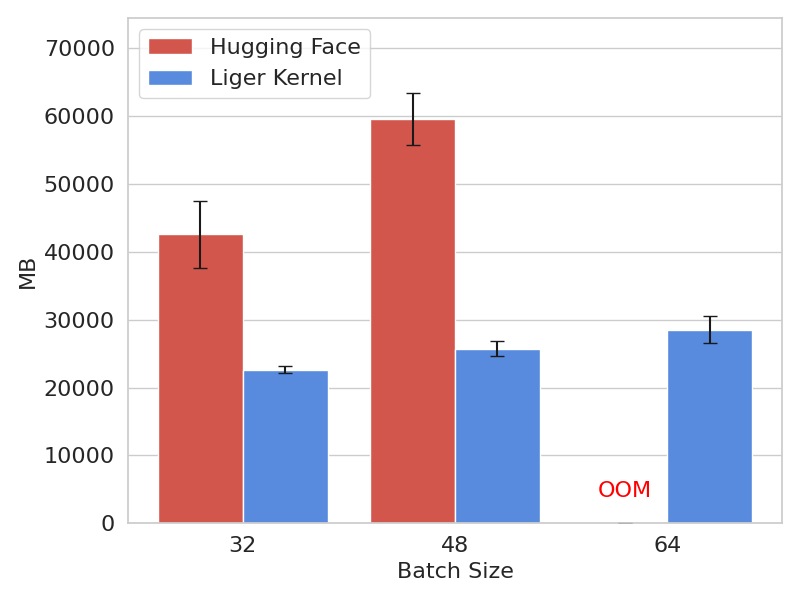}
        \label{fig:memory_Qwen2}
    \end{minipage}\hfill
    \begin{minipage}{0.45\textwidth}
        \centering
        \includegraphics[width=\textwidth]{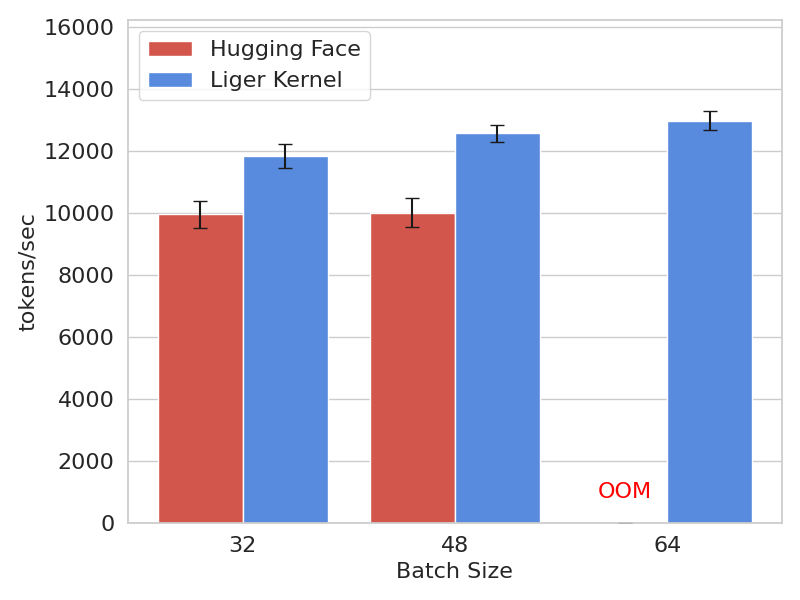}
        \label{fig:throughput_Qwen2}
    \end{minipage}
    \caption{Comparison of peak allocated memory and throughput for Qwen2.}
    \label{fig:combined_Qwen2}
\end{figure}

\begin{figure}[H]
    \centering
    \begin{minipage}{0.45\textwidth}
        \centering
        \includegraphics[width=\textwidth]{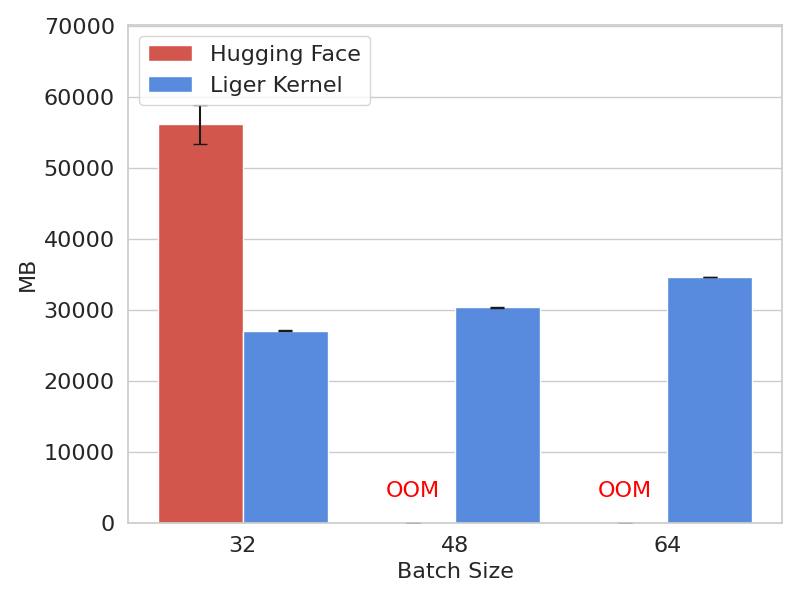}
        \label{fig:memory_Gemma_7b}
    \end{minipage}\hfill
    \begin{minipage}{0.45\textwidth}
        \centering
        \includegraphics[width=\textwidth]{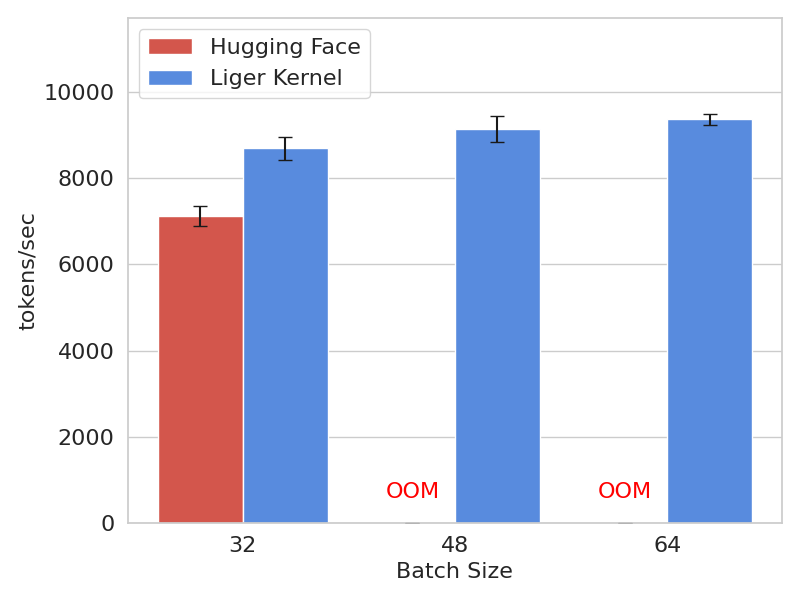}
        \label{fig:throughput_Gemma_7b}
    \end{minipage}
    \caption{Comparison of peak allocated memory and throughput for Gemma 7b.}
    \label{fig:combined_Gemma}
\end{figure}

\begin{figure}[H]
    \centering
    \begin{minipage}{0.45\textwidth}
        \centering
        \includegraphics[width=\textwidth]{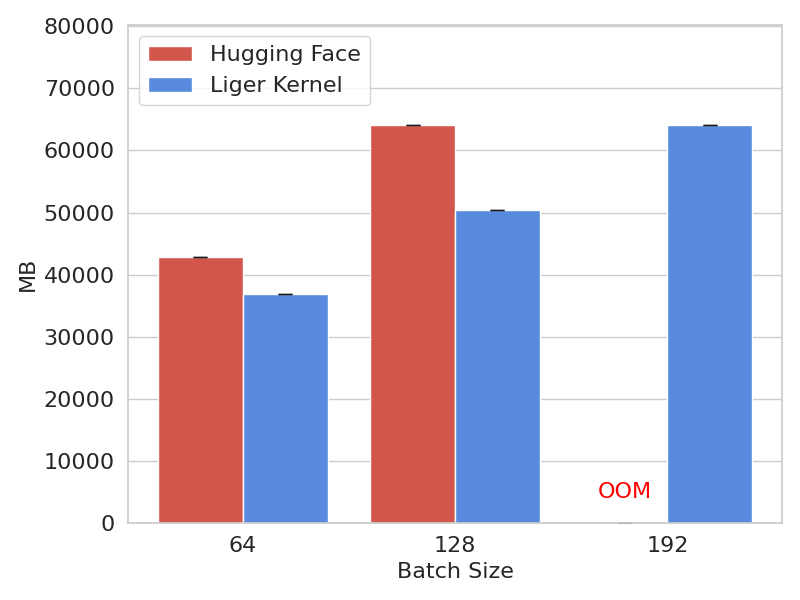}
        \label{fig:memory_mistral_7b}
    \end{minipage}\hfill
    \begin{minipage}{0.45\textwidth}
        \centering
        \includegraphics[width=\textwidth]{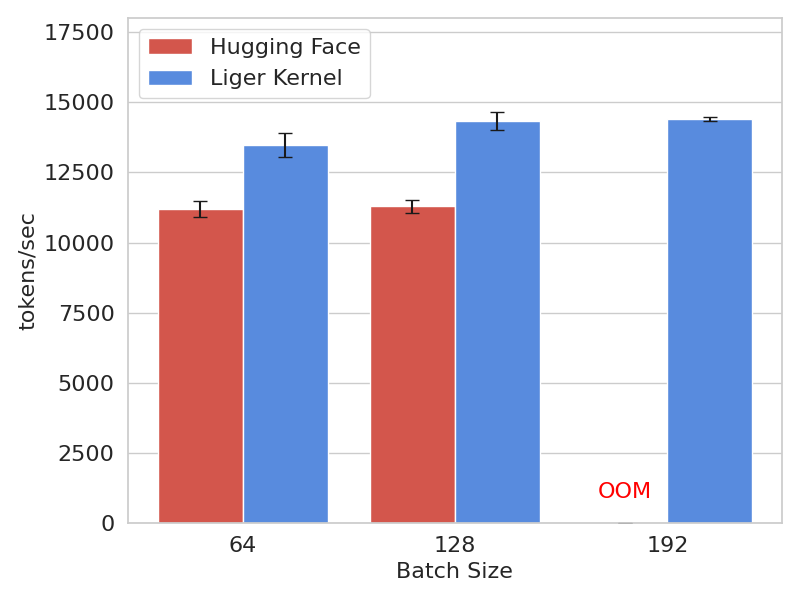}
        \label{fig:throughput_mistral_7b}
    \end{minipage}
    \caption{Comparison of peak allocated memory and throughput for Mistral 7b.}
    \label{fig:combined_mistral}
\end{figure}

\begin{figure}[H]
    \centering
    \begin{minipage}{0.45\textwidth}
        \centering
        \includegraphics[width=\textwidth]{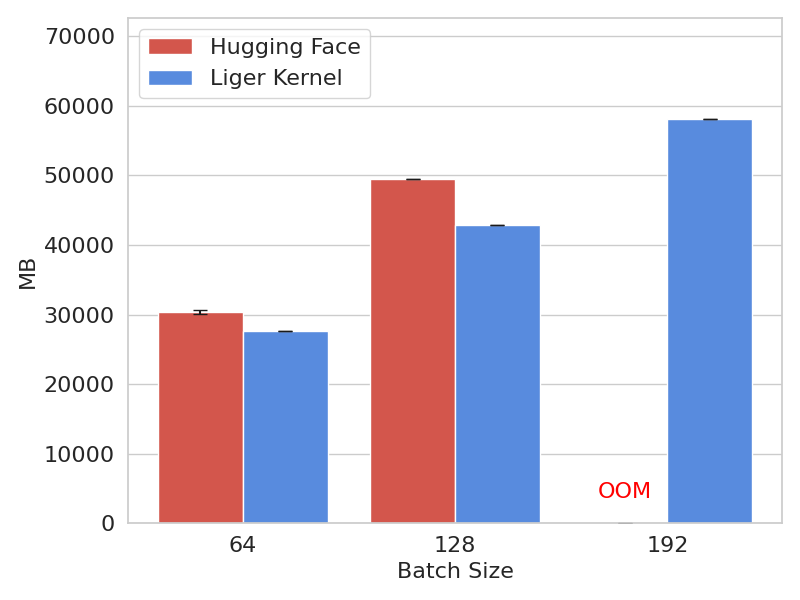}
        \label{fig:memory_phi3}
    \end{minipage}\hfill
    \begin{minipage}{0.45\textwidth}
        \centering
        \includegraphics[width=\textwidth]{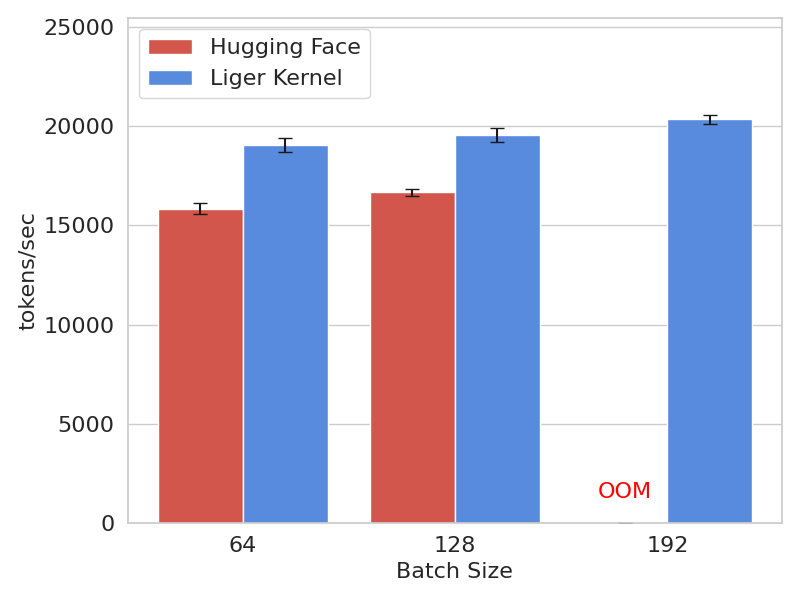}
        \label{fig:throughput_phi3}
    \end{minipage}
    \caption{Comparison of peak allocated memory and throughput for Phi3.}
    \label{fig:combined_phi3}
\end{figure}

At a batch size of $64$, LLaMA 3-8B demonstrates a \textbf{42.8\% increase in throughput}, coupled with a \textbf{54.8\% reduction in GPU memory usage} (Figure \ref{fig:combined_llama3_8b}). This enables training on smaller GPUs or using larger batch sizes and longer sequence lengths with lower resource consumption. Similarly, at a batch size of $48$ our kernels improve the throughput of Qwen2 by \textbf{25.5\%}, while achieving a \textbf{56.8\% reduction in GPU memory usage} (Figure \ref{fig:combined_Qwen2}). 
For Gemma, throughput improves by \textbf{11.9\%} with a \textbf{51.8\% reduction in memory usage} at a batch size of $48$ (Figure \ref{fig:combined_Gemma}). Mistral, at a batch size of $128$, exhibits a \textbf{27\% increase in throughput}, with a \textbf{21\% drop in GPU memory usage} (Figure \ref{fig:combined_mistral}). Finally, Phi3, at a batch size of $128$, shows a \textbf{17\% increase in throughput}, while reducing memory usage by \textbf{13\%} (Figure \ref{fig:combined_phi3}).  Overall, the results highlight several notable use cases. LLaMA 3-8B's exceptional improvements make it ideal for resource-constrained environments where GPU memory is a bottleneck. Additionally, Qwen2's strong memory reductions position it well for tasks involving large datasets or extended training durations. Mistral's high throughput gains make it advantageous for workloads requiring large batch sizes.

\paragraph{Medusa.}

Medusa~\citep{medusa2024} is a simple framework that democratizes acceleration techniques for LLM generation by using multiple decoding heads to predict several subsequent tokens in parallel. During training, Medusa requires adding \(k\) decoding heads to the hidden states right before the regular LM head \(h_t\). The \(k\)-th head is used to predict the token in the \((t + k + 1)\)-th position of the next tokens (the original language model head is used to predict the \((t + 1)\)-th position).

The Liger LFCE kernel is particularly effective in this context, as it eliminates the need to materialize logits for each decoding head. This is critical in scenarios with large vocabulary sizes, such as LLaMA-3's 128k tokens, where materializing logits can lead to significant memory consumption. The introduction of multiple decoding heads often results in out of memory issues. However, by leveraging the Liger fused CE kernel, which computes gradients in place without materializing logits, we achieve highly efficient results. This approach enables further exploration and development in multi-token prediction.

\begin{figure}[H]
    \centering
    \begin{minipage}{0.45\textwidth}
        \centering
        \includegraphics[width=\textwidth]{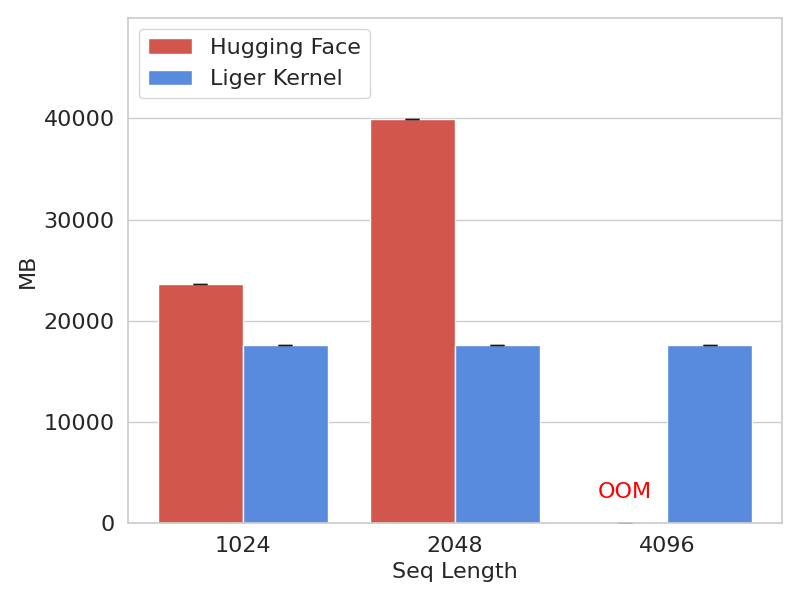}
        \label{fig:memory_stage1_head3}
    \end{minipage}\hfill
    \begin{minipage}{0.45\textwidth}
        \centering
        \includegraphics[width=\textwidth]{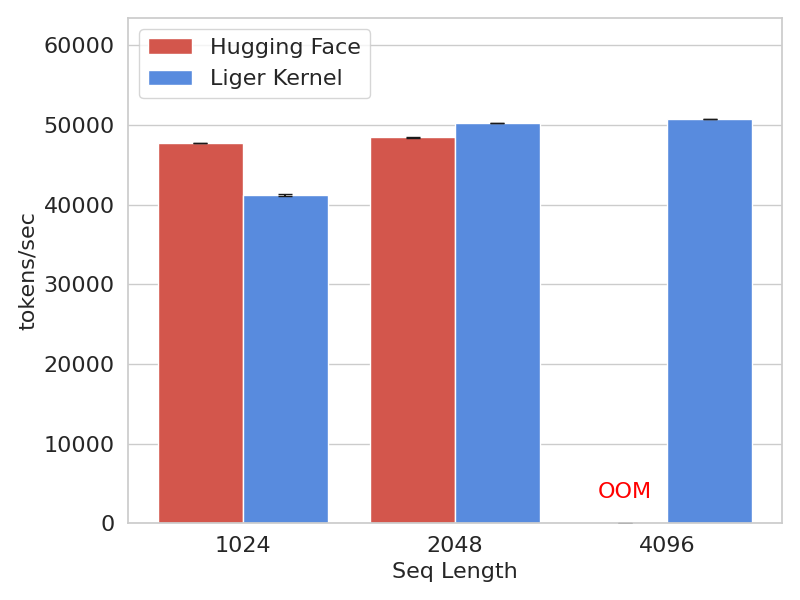}
        \label{fig:throughput_stage1_head3}
    \end{minipage}
    \caption{Comparison of peak allocated memory and throughput for Stage 1 with 3 Medusa heads.}
    \label{fig:combined_stage1_head3}
\end{figure}

\begin{figure}[H]
    \centering
    \begin{minipage}{0.45\textwidth}
        \centering
        \includegraphics[width=\textwidth]{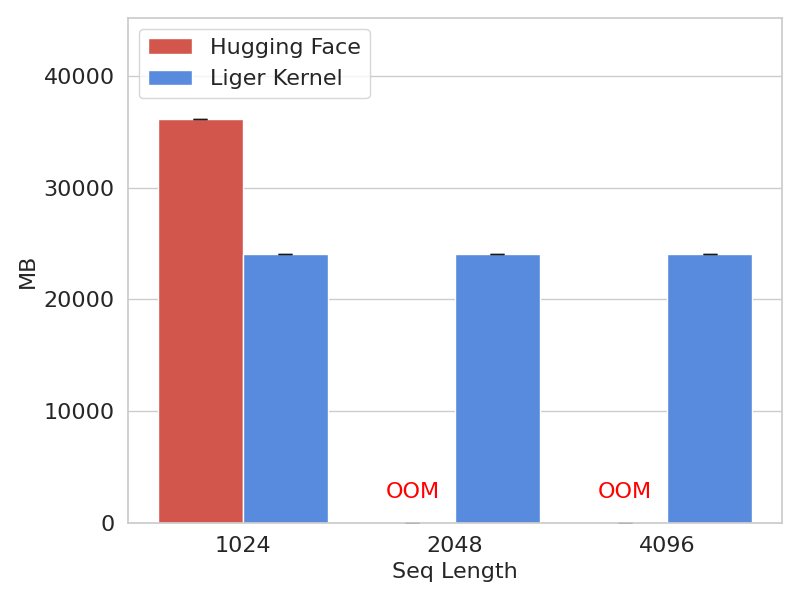}
        \label{fig:memory_stage1_head5}
    \end{minipage}\hfill
    \begin{minipage}{0.45\textwidth}
        \centering
        \includegraphics[width=\textwidth]{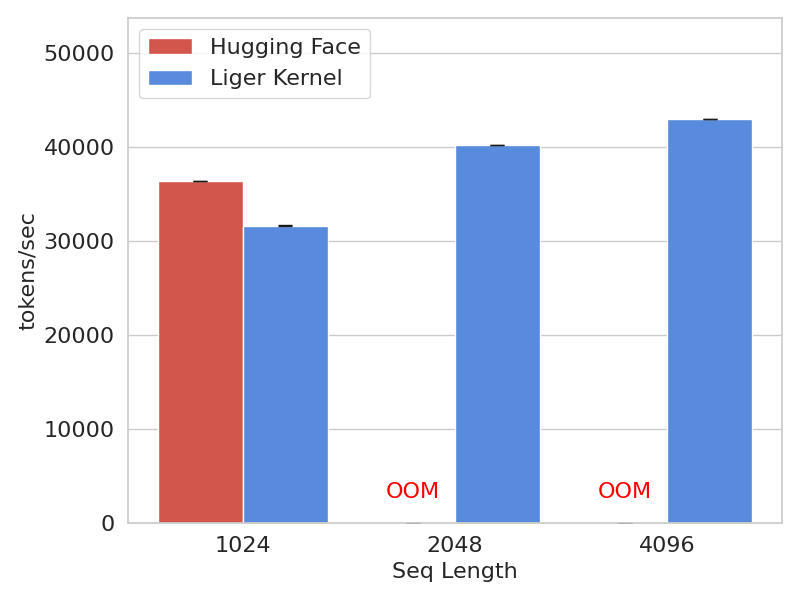}
        \label{fig:throughput_stage1_head5}
    \end{minipage}
    \caption{Comparison of peak allocated memory and throughput for Stage 1 with 5 Medusa heads.}
    \label{fig:combined_stage1_head5}
\end{figure}

\begin{figure}[H]
    \centering
    \begin{minipage}{0.45\textwidth}
        \centering
        \includegraphics[width=\textwidth]{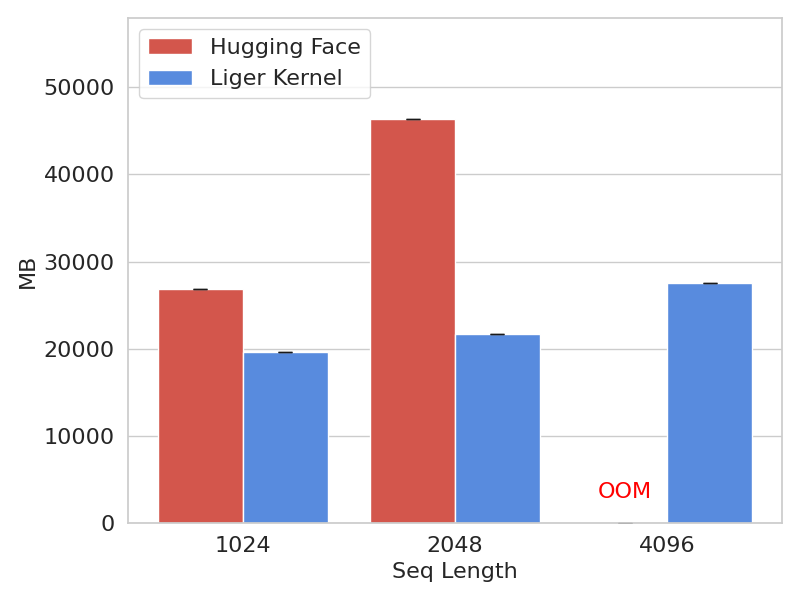}
        \label{fig:memory_stage2_head3}
    \end{minipage}\hfill
    \begin{minipage}{0.45\textwidth}
        \centering
        \includegraphics[width=\textwidth]{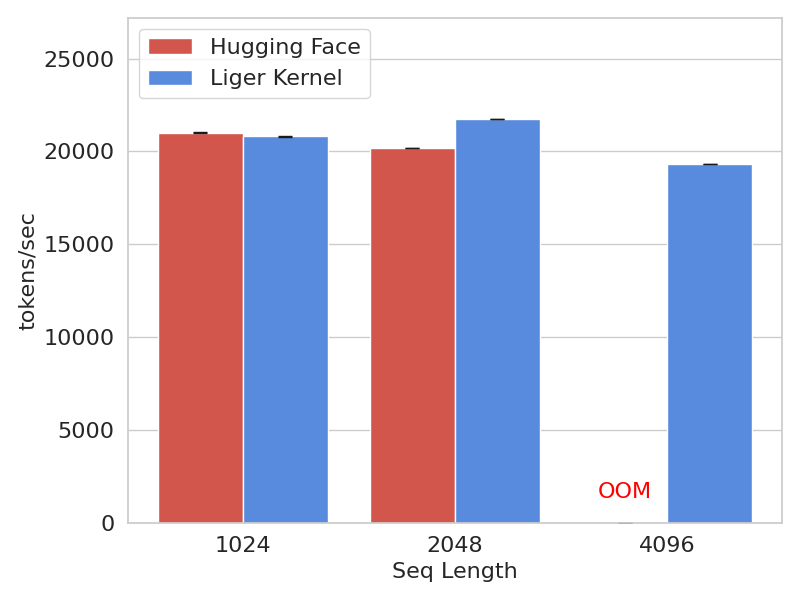}
        \label{fig:throughput_stage2_head3}
    \end{minipage}
    \caption{Comparison of peak allocated memory and throughput for Stage 2 with 3 Medusa heads.}
    \label{fig:combined_stage2_head3}
\end{figure}

\begin{figure}[H]
    \centering
    \begin{minipage}{0.45\textwidth}
        \centering
        \includegraphics[width=\textwidth]{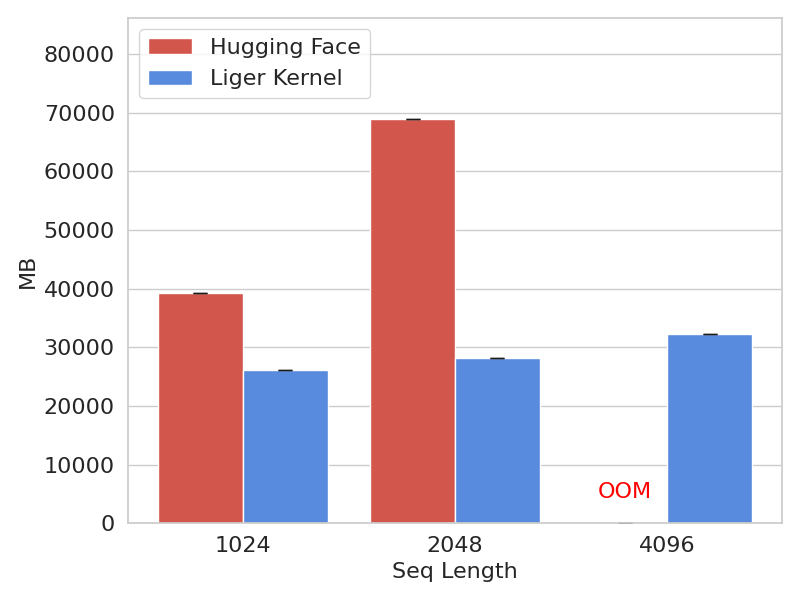}
        \label{fig:memory_stage2_head5}
    \end{minipage}\hfill
    \begin{minipage}{0.45\textwidth}
        \centering
        \includegraphics[width=\textwidth]{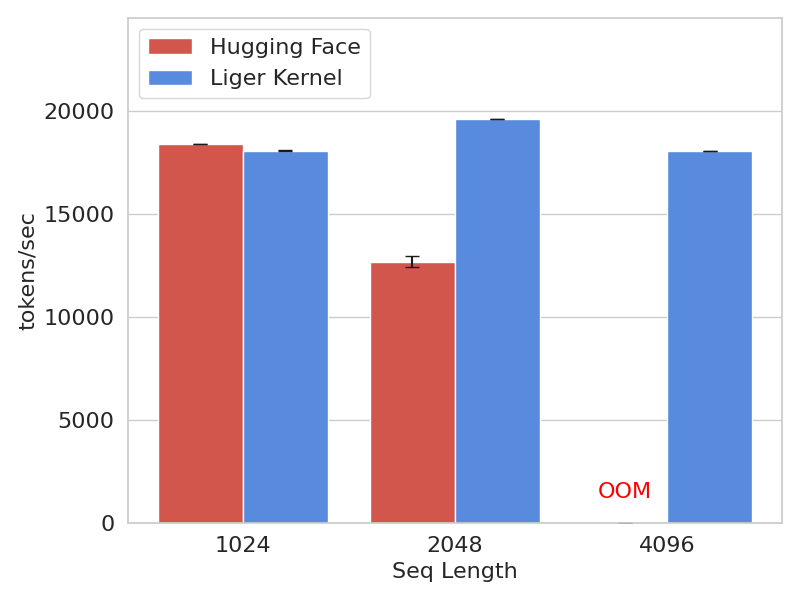}
        \label{fig:throughput_stage2_head5}
    \end{minipage}
    \caption{Comparison of peak allocated memory and throughput for Stage 2 with 5 Medusa heads.}
    \label{fig:combined_stage2_head5}
\end{figure}

Medusa training has two flavors. The first, called stage-1, involves training only the additional Medusa heads while keeping the backbone LLM frozen. The second approach tunes both the backbone and the LLM heads simultaneously. We have benchmarked both cases, and the Liger kernel has demonstrated reduced memory usage and improved throughput. Without the Liger kernel, experiments are highly prone to out of memory issues. In Figures \ref{fig:combined_stage1_head3}-\ref{fig:combined_stage2_head5}, the standard errors measured from repetitive runs are typically less than $1\%$ hence not visible from most of the plots.

\noindent\textbf{Note:} This technical report focuses solely on performance benchmarking. Generating effective LLM heads that can accelerate inference for the LLaMA3-8B model is not within the scope of this report. Such work requires extra work for training data selection, hyperparameter tuning, and warmup techniques to ensure proper model convergence. Our experiments utilize $8$ NVIDIA A100 GPUs ($80$ GB each) to train the LLaMA 3-8B model with a variable sequence length, a batch size of $4$, bfloat16 precision and the AdamW optimizer.

\section{Conclusions}

Liger Kernel offers optimized Triton kernels that improve training efficiency with a user-friendly API, seamless integration with popular frameworks, and a commitment to performance. Our goal is to make Liger Kernel the leading open-source Triton kernel library for LLM training. We aim to achieve this by focusing on:

\begin{itemize}
    \item \textbf{Ease of Use:} Offering intuitive APIs, broad model support, and wide hardware compatibility
    \item \textbf{Performance Focus:} Maximizing computational efficiency and ensuring exactness.
    \item \textbf{Ecosystem Engagement:} Building a strong community through events and collaborations with industry leaders, alongside fostering recognition and branding for contributors.
    \item \textbf{Operational Excellence:} Ensuring stable CI, rigorous testing protocols, and an active community.
\end{itemize}

With these commitments, \texttt{Liger-Kernel} aspires to become the preferred choice for efficient and scalable LLM training, driving innovation and adoption within the deep learning community. While existing work primarily focuses on training, the same techniques can be seamlessly adapted for optimizing model inference.

\section{Contributors and Acknowledgements}

\subsection{Core Contributors}

\textbf{Pin-Lun Hsu} Project lead. Led, architected, and implemented multiple kernels, public interface, and test suite.

\noindent \textbf{Yun Dai} Core contributor. Designed an efficient version of RoPE, GeGLU, and improved the precision of Fused Linear CrossEntropy. Designed the public interface.

\noindent \textbf{Vignesh Kothapalli} Core contributor. Implemented Fused Linear CrossEntropy and designed the scaling and sharding formula.

\noindent \textbf{Qingquan Song} Core contributor. Implemented SwiGLU. Led the convergence tests and PyTorch lightning integration. Ensure the contiguity of RoPE and kernel testing precisions.

\noindent \textbf{Shao Tang} Core contributor. Implemented Layer Norm variants. Derived gradient formulas for different cases. Proposed best kernel practices, including ensuring contiguity and conducting convergence tests.

\noindent \textbf{Siyu Zhu} Core contributor. Implemented Fused Linear CrossEntropy and adapted the kernel for the Medusa (multi-token prediction) use case, proving its effectiveness with benchmarks. Led the Hugging Face integration.

\noindent \textbf{Steven Shimizu} Contributor. Improved HuggingFace integration and contributed to the tests.

\noindent \textbf{Shivam Sahni} Contributor. Expanded model support and made several kernel improvements.

\noindent \textbf{Haowen Ning} Contributor and the overall team lead of LLM training infra.

\noindent \textbf{Yanning Chen} Contributor and the team manager.

\subsection{Acknowledgement} 
We thank AMD and Intel for funding GPUs for our AMD and Intel CI. We also thank Modal for funding 3000 credits from GPU MODE IRL for our NVIDIA CI.

We thank Triton\footnote{\href{https://triton-lang.org/main/getting-started/tutorials/index.html}{https://triton-lang.org/main/getting-started/tutorials/index.html}}, flash-attention\footnote{\href{https://github.com/dao-ailab/flash-attention}{https://github.com/dao-ailab/flash-attention}}, and Unsloth\footnote{\href{https://github.com/unslothai/unsloth}{https://github.com/unslothai/unsloth}} for the reference of Triton kernels for LLM training, tiny shakespeare dataset\footnote{\href{https://huggingface.co/datasets/karpathy/tiny_shakespeare}{https://huggingface.co/datasets/karpathy/tiny\_shakespeare}}  and llm.c\footnote{\href{https://github.com/karpathy/llm.c}{https://github.com/karpathy/llm.c}} for convergence testing design, Efficient Cross Entropy\footnote{\href{https://github.com/mgmalek/efficient_cross_entropy}{https://github.com/mgmalek/efficient\_cross\_entropy}} for fused linear cross entropy reference, AutoAWQ\footnote{\href{https://github.com/casper-hansen/AutoAWQ}{https://github.com/casper-hansen/AutoAWQ}} and \textbf{Robert Shaw} for Automodel design, as well as Hugging Face, PyTorch Lightning, Axolotl, and Llama-Factory for the collaboration.

We also thank our leaders \textbf{Animesh Singh} and \textbf{Kapil Surlaker} for their invaluable expertise in the ML infrastructure stack and open-source strategy.

Thanks to \textbf{Claire (Yi-Shan) Wu} for the LOGO design and Wave Snippets\footnote{\href{https://www.wavesnippets.com/}{https://www.wavesnippets.com/}} for generating the animated code snippets.

\bibliographystyle{plainnat} 
\bibliography{ref} 
\end{document}